\documentclass{article}

\usepackage{microtype}
\usepackage{graphicx}
\usepackage{booktabs} 
\usepackage[font=small,skip=-2pt]{caption}

\usepackage{hyperref}

\usepackage{subcaption}

\usepackage[accepted]{icml2021}

\begin{document}

\twocolumn[
\icmltitle{Using a Cross-Task Grid of Linear Probes to Interpret CNN Model Predictions On Retinal Images}




\begin{icmlauthorlist}
\icmlauthor{Katy Blumer}{cornell,goo}
\icmlauthor{Subhashini Venugopalan}{goo}
\icmlauthor{Michael P. Brenner}{goo,har}
\icmlauthor{Jon Kleinberg}{cornell}
\end{icmlauthorlist}

\icmlaffiliation{cornell}{Cornell University}
\icmlaffiliation{goo}{Google Research}
\icmlaffiliation{har}{Harvard University}

\icmlcorrespondingauthor{KB}{kblumer@google.com}
\icmlcorrespondingauthor{SV}{vsubhashini@google.com}

\icmlkeywords{Machine Learning, ICML}

\vskip 0.15in
]



\printAffiliationsAndNotice{}

\begin{abstract}
We analyze a dataset of retinal images using linear probes: linear regression models trained on some ``target'' task, using embeddings from a deep convolutional (CNN) model trained on some ``source'' task as input. We use this method across all possible pairings of 93 tasks in the UK Biobank dataset of retinal images, leading to $\sim$164k different models. We analyze the performance of these linear probes by source and target task and by layer depth. 
\\
We observe that representations from the middle layers of the network are more generalizable. 
We find that some target tasks are easily predicted irrespective of the source task, and that some other target tasks are more accurately predicted from correlated source tasks than from embeddings trained on the same task.
\end{abstract}

\section{Introduction}
\label{intro}
Retinal fundus (internal eye) images have been well-studied in machine learning applications. Machine learning can predict retinal disease with great accuracy \cite{gulshan2016development,badar2020application}. However, many other, often surprising, features can also be predicted from these images: for example, visual acuity~\cite{varadarajan2018deep}, cardiovascular risk~\cite{poplin2018prediction}, diabetes~\cite{zhang2021deep}, anaemia~\cite{mitani2020detection} and many other variables \cite{rim2020prediction}. Many of these are novel predictions not known to be predictable from these images by human experts, and it would be useful to understand precisely which features in the fundus image make these features predictable. 
   
The challenge in even framing this question is the lack of tractable formalisms for characterizing \textit{how} predictions are made. One simple idea is to take a model that achieves the surprising outcome of predicting certain variables from retinal images and ask what else this model is able to predict effectively. 
To do this, we need a way to evaluate how effectively we can build simple extensions of the models' internal representation to predict other quantities of interest. 
Linear probes \cite{alain2016understanding} and concept activation vectors (TCAV) \cite{kim2018interpretability} are techniques that provide paths to doing this; in this work we focus on linear probes.
\\\\
Linear probes have been widely used for interpretability to understand performance of deep models with application to language processing~\cite{hewitt2019designing,hewitt2019structural,belinkov2021probing}, computer vision~\cite{alain2016understanding,asano2019self}, speech~\cite{oord2018representation} or generally  when understanding different neural network architectures~\cite{raghu2017svcca, graziani2019interpreting, horoi2020internal}.
In this paper, we use this technique to study and understand CNN model predictions on 93 different tasks based on the UK Biobank dataset~\cite{sudlow2015uk} of retinal images and labels.

We find that embeddings from the middle layers of the networks (as opposed to those closest to the output) learn features that are more generalizable across multiple target tasks, with linear probes consistently making accurate predictions. Additionally, we find that some target tasks such as eye position (left vs. right eye) and refractive error are easily predicted irrespective of the source task the CNN was trained on. We also find that other target tasks such as height are better predicted by embeddings trained on correlated tasks such as blood testosterone or self-reported sex, than by embeddings trained on the original task.  Ultimately these results give insight into which features of the input data make it possible to learn different target values.

\section{Methods}
\label{methods}
\begin{figure*}[htb]
    \centering
    \includegraphics[width=\textwidth]{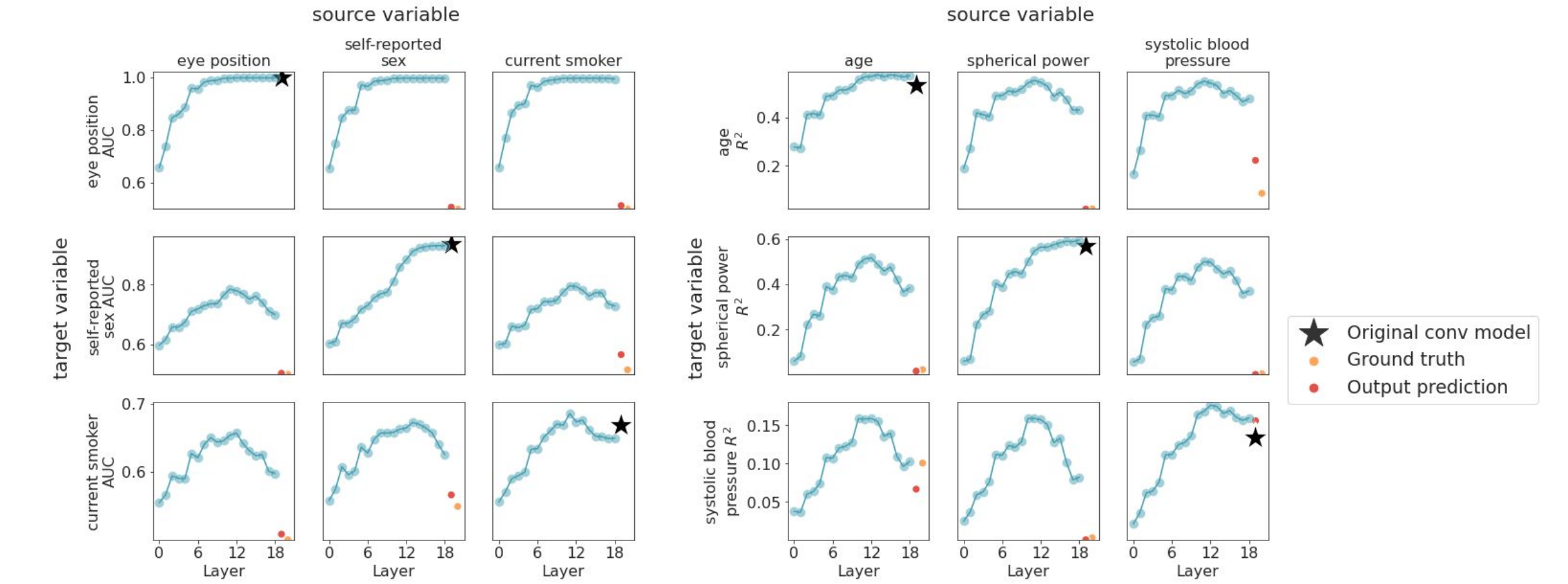}
    \caption{Plots of linear model performance by layer, organized by target variable (row) and source variable (column). X-axis scale is consistent along each row. Orange and red dots are from linear models trained on the single-valued source prediction and source variable value, respectively.} 
    \label{fig:shape}
\end{figure*}

\begin{figure*}[hbt]
    \centering
    \includegraphics[width=0.83\textwidth]{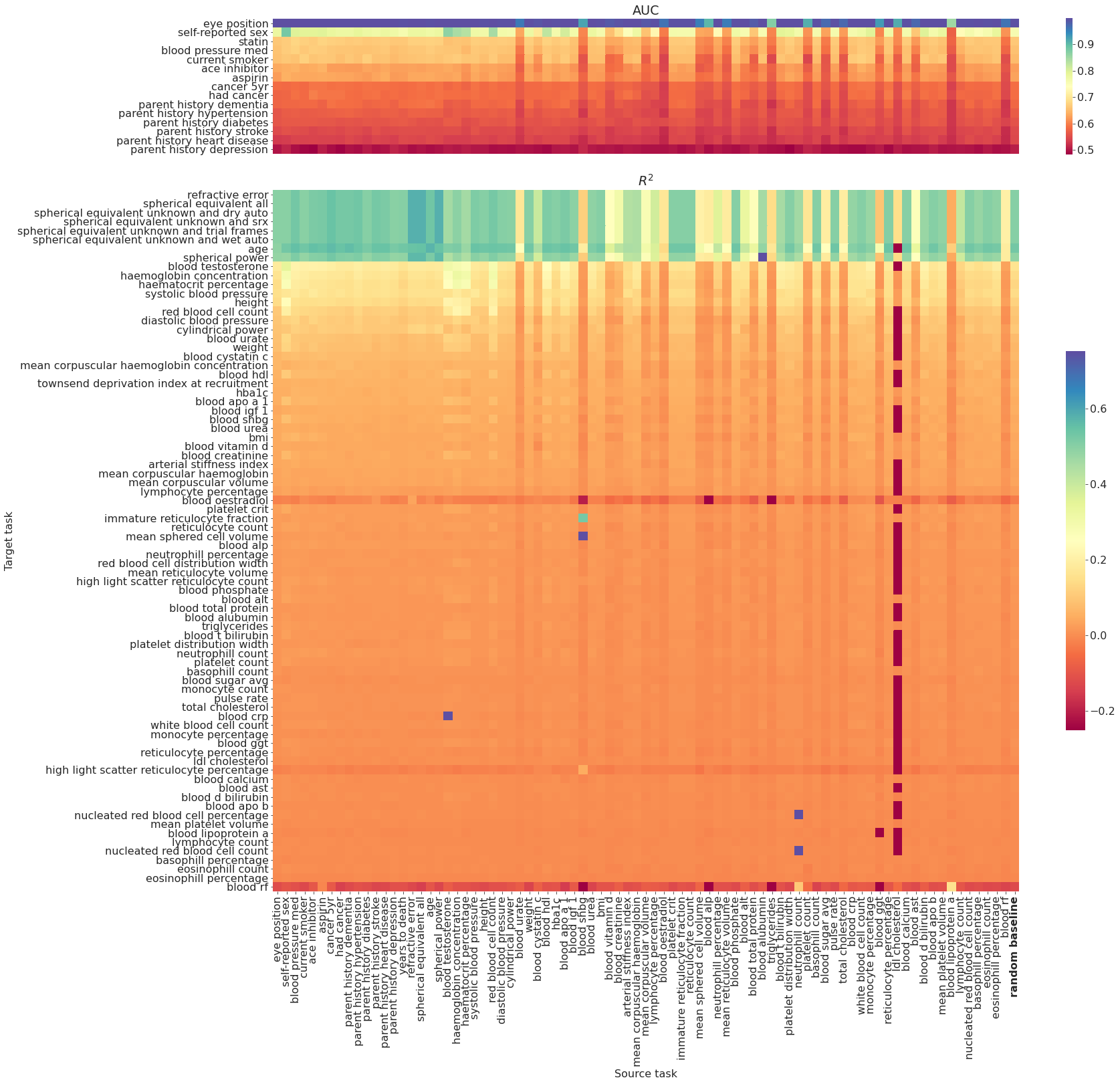}
    \caption{Cross-comparison of all tasks for layer 12. This layer was chosen as it was close to the best layer for most models, and showed the most interesting variation between variables. Tasks are ordered first by whether they're binary or continuous values, then by performance of that task as a target with the ``eye position'' source task (in practice, this worked almost as well as hierarchical clustering). The bolded source task is a random baseline, where the source CNN was trained on labels drawn from the standard uniform distribution. 
    }
    \label{fig:heatmap}
\end{figure*}

\paragraph{Data}
\label{data}
We used a dataset of retinal images from the UK Biobank study \cite{sudlow2015uk} containing 140,000 retinal fundus images from 68,000 patients. We separated 12.5\% of the patients into a test set, with the rest  in the train set. We used a set of 93 non-eye-disease-related variables available in the UK Biobank resource. These included demographic data such as age and self-reported sex (recorded in the data as female/male); measurements such as blood tests and visual acuity; and miscellaneous features such as eye position (whether the image is of the left or right eye). Figure \ref{fig:heatmap} summarizes all 93 variables. 

\subsection{Experiments}
\label{experiments}
We trained identical deep convolutional Inception V3 models for each of the 93 variables in our dataset. The models were pre-trained on ImageNet with auxiliary loss turned off and then trained with early stopping for a maximum of 200,000 steps.
We then trained linear regression models for the same set of variables. Each model used the output of an intermediate layer from one of the convolutional models as input. We used 19 different intermediate layers spanning the depth of the Inception V3 architecture, and spatially average-pooled the output to make an exact linear regression tractable. These models were trained on the same training set as the convolutional models. 

In total, this gave rise to $\sim$164k different linear models: 93 ``source'' tasks (convolutional models) $\times$ 93 ``target'' tasks (linear models) $\times$ 19 intermediate layers. We evaluated each linear model on the test dataset and calculated either an AUC or coefficient of determination ($R^2$), depending on whether the target task/variable was binary or continuous. 

Along with linear regressions on intermediate layers, we also carried out linear regressions on the raw values of the variables themselves and on the single-valued predictions from each convolutional model, in order to distinguish tasks that shared common representations from tasks that were merely correlated with each other. The regressions were done on the training set and evaluated on the test set, just like the intermediate layer regressions.

\section{Analysis and observations}
In this section we present our analysis and observations from our experiments.

\subsection{Middle layer representations generalize best}
Figure \ref{fig:shape} presents performance of the linear probe models (AUC or $R^2$ on the y axis) on embeddings across layers (x axis) from models trained to predict different variables (self-reported sex, eye position, smoker status, etc.) represented in each column. We observe that performance across layers tends to follow the same pattern: it increases from the layers closest to the input until the middle layers, and then decreases again - except where the source and target tasks are the same (on-diagonal plots). When the source and target are the same, the performance plateaus or continues to increase in the final layers. 

The shape of the graph looks more similar along rows than columns, suggesting that the difficulty of learning a given task is more important than the differences between input embeddings learned for different tasks. This observation is in contrast with typical transfer learning setting where it is much more common to tune the final layers of a model (layers closer to the output) for new tasks, suggesting that we may want to entirely choose a layer in the middle for transfer learning.

In order to distinguish embeddings with similar representations from mere variable correlation, we also trained linear regression models on each input task's ground truth values and the convolutional model's output predictions (Fig \ref{fig:shape}, red and orange points). These generally do much worse than models trained on intermediate embeddings, though there are cases where the performance is comparable. One such example is predicting systolic blood pressure with age as the source task, which makes sense as the two variables are known to be correlated \cite{hypertension03}.

\subsection{Specific middle layers perform best, generalizing well across multiple tasks}
\begin{figure}[h]
    \centering
    \includegraphics[width=\linewidth]{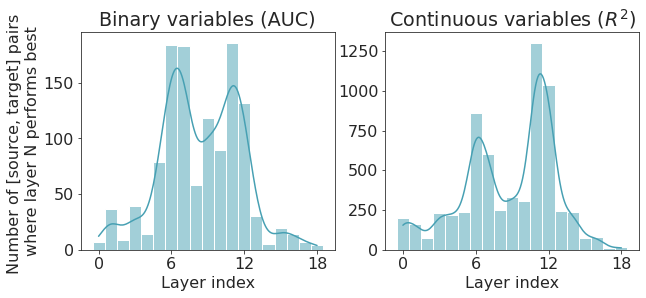}
    \caption{Histograms of best-performing layers for each [source, target] variable pair. Pairs where the two variables were the same are excluded.}
    \label{fig:histograms}
\end{figure}
\begin{figure}[h]
    \centering
    \includegraphics[width=0.5\linewidth]{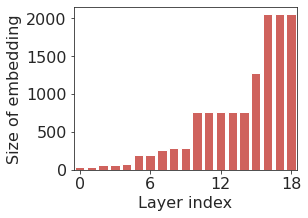}
    \caption{\small{Dimensions of embeddings (input to linear regression models) by layer of the source convolutional model. Note that embeddings were spatially average-pooled, so the dimension is just the number of channels in that layer's output.}}
    \label{fig:layer_size}
\end{figure}
Next we ask, for the Inception network are there specific layers that give the best performance when generalizing to all tasks? Figure \ref{fig:histograms} shows a histogram of the best-performing layer for each [source, target] variable pair. (Pairs where the source and target tasks are the same were excluded.) 
It is interesting to see two distinct peaks in the middle layers (around layers 6,7 as well as layers 11, 12). However layer 11 appears to be consistently more generalizable and is amongst the top 3 layers with best performance on many of our pairwise comparisons. We don't know, however, why there are \textit{two} peaks. The linear models' input does have different dimensions depending on the layer (Fig \ref{fig:layer_size}), but the peaks don't obviously correspond to the changes in size. The clustering could also be due to correlations between [source, target] pairs. This would be interesting to investigate further.

\subsection{Performance bands for same target task}
\begin{figure}
    \centering
    \includegraphics[width=\linewidth]{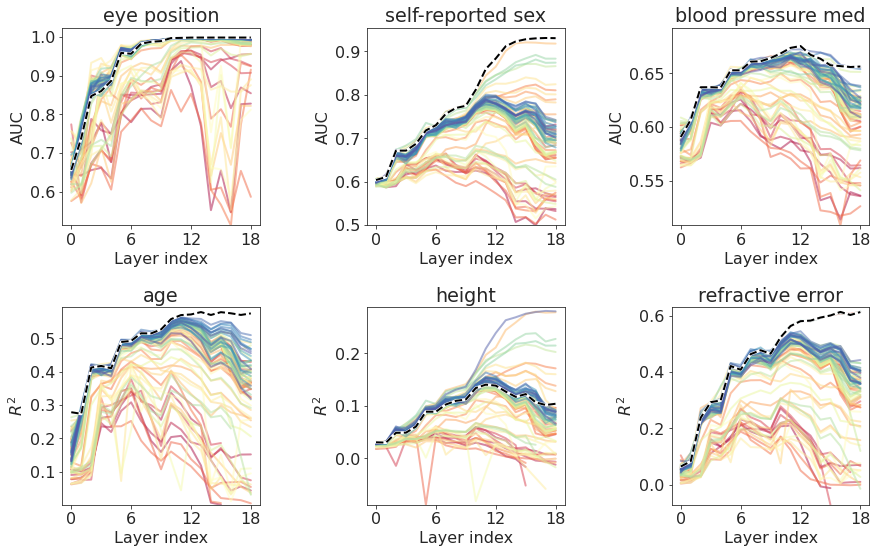}
    \caption{Comparison of performance on all source tasks for a given target task. Each line represents one source task, and the color is based on the performance of that variable as a \textit{target} task on itself as a source task. Blue is best performance, red is worst. Dotted lines show the performance when the source variable is the same as the target. Observe that for `height' representations from other source tasks (in this case, self-reported sex and blood testosterone) are better predictors.}
    \label{fig:bands}
\end{figure}
As we would expect, the models generally perform best when the source and target tasks are the same (dotted lines, Fig \ref{fig:bands}). This is not always true, however - for example. ``height'' performs better on several other source tasks than on itself as a source task. The other sources were tasks such as as blood testosterone and self-reported sex, which share two properties: they're correlated with height, and are easier to predict than height (for which we never get an $R^2$ above ~0.3). We might guess that, because height is so hard to predict, there's not as much room for embeddings trained on it to improve - an illustration of the utility of multitask learning.

We can also observe that there are clear bands and outliers. The bands are closer together in the earlier layers, likely because features learned in the earlier layers are more universal. The outliers generally make sense: for example, the ``blood pressure medication'' as target task performs well with itself as a source task, but it performs just as well on source tasks for two related medications, aspirin and ACE inhibitors.

\subsection{Comparison across all tasks}
When comparing a single layer for all task combinations, we again see that the target task is usually more important than the source task. But there are some other interesting relationships - for example, age, eye position (left vs. right eye), and most measures of visual acuity are easily predicted regardless of the source task. Age is correlated with a small subset of tasks (see supplement), but predicting age with those as source tasks gets only about the same performance as many other source task.



\section{Conclusion}
To conclude, in this work we used the basic notion of linear probes to study models trained on 93 different tasks/variables in fundus images in the UK Biobank. 
We looked at nearly 164k models, examining  different variables as source and target pairs. We find interesting patterns in performance on source embeddings at various layer depths, and in performance on various combinations of source and target task.

Overall, the results show that simple linear probes provide a rich environment for unravelling the relationships between the underlying data and labels, providing insight into why neural networks trained on single labels are able to make accurate predictions. Future work will use the different representations to unravel which features of images are responsible for the different accurate predictions.
\paragraph{Acknowledgements}
This research has been conducted using the UK Biobank Resource \cite{sudlow2015uk}. We would like to thank Abbi Ward for help with permissions and access to the dataset, Avinash Varadarajan for help with the dataset, annotations, and support with the training and evaluation infrastructure, Xinle Sheila Lu for helpful pointers on model inference, and Yun Liu and Maithra Raghu for helpful conversations and references.

\clearpage


\bibliography{example_paper}
\bibliographystyle{icml2021}



\end{document}